%% file: root.tex
\documentclass[letterpaper, 10 pt, conference]{ieeeconf}  

\IEEEoverridecommandlockouts                              

\usepackage{graphics} 
\usepackage{amsmath,amssymb}
\usepackage{graphicx} 
\usepackage{placeins}

\usepackage{algorithm} 
\usepackage{algpseudocode}
\usepackage{multicol}
\usepackage{hyperref}
\usepackage{url}
\usepackage{multirow}
\usepackage{adjustbox}
\usepackage{etoolbox}
\usepackage{paralist}
\usepackage{flushend}
\usepackage[symbol]{footmisc}
\usepackage{afterpage}
\usepackage[font=small]{caption}
\usepackage{float}
\usepackage{paralist}
\usepackage{flushend}
\usepackage[symbol]{footmisc}

\title{\LARGE \bf
SplatSim: Zero-Shot Sim2Real Transfer of RGB Manipulation Policies Using Gaussian Splatting
}

\author{M. Nomaan Qureshi$^{1}$, Sparsh Garg$^{1}$, Francisco Yandun$^{1}$, David Held$^{1}$, George Kantor$^{1}$, Abhisesh Silwal$^{1}$
\thanks{{1} All authors are with the Carnegie Mellon University, USA.}
}

\begin{document}

\maketitle

\thispagestyle{empty}
\pagestyle{empty}


\begin{abstract}
Sim2Real transfer, particularly for manipulation policies relying on RGB images, remains a critical challenge in robotics due to the significant domain shift between synthetic and real-world visual data. In this paper, we propose \textit{SplatSim}, a novel framework that leverages Gaussian Splatting as the primary rendering primitive to reduce the Sim2Real gap for RGB-based manipulation policies. By replacing traditional mesh representations with Gaussian Splats in simulators, \textit{SplatSim} produces highly photorealistic synthetic data while maintaining the scalability and cost-efficiency of simulation. We demonstrate the effectiveness of our framework by training manipulation policies within \textit{SplatSim} and deploying them in the real world in a zero-shot manner, achieving an average success rate of 86.25\%, compared to 97.5\% for policies trained on real-world data. Videos can be found on our project page: \href{https://splatsim.github.io}{https://splatsim.github.io}

\end{abstract}

\input{sections/1_intro}
\input{sections/2_related_works}
\input{sections/3_method}

\input{sections/4_experiments}
\input{sections/5_conclusion}




\section*{Acknowledgement}
We would like to express our gratitude to Prof. Shubham Tulsiani for his valuable insights during the early development of this idea. We also extend our thanks to Anurag Ghosh and Ayush Jain for their thoughtful feedback in refining the manuscript.
This work is in part supported by NSF/USDA-NIFA AIIRA AI Research Institute 2021-67021-35329 and USDA-NIFA/NSF National Robotics Initiative 2021-67021-35974.



\bibliographystyle{IEEEtran}
\bibliography{root}



\end{document}

%% file: sections/1_intro.tex
\section{INTRODUCTION}


 The Sim2Real problem, a focal challenge in robotics, pertains to the transfer of control policies learned in simulated environments to real world settings. Recently, significant progress has been made in deploying controllers trained in simulation to the real world in a zero-shot manner. Robots have demonstrated the ability to walk on rough terrains~\cite{agarwal2022legged, 10610200}, perform in-hand object rotation~\cite{qi2022inhand, 10610532, dextreme, YinHQCW23}, and grasp previously unseen objects~\cite{lum2024dextrahgpixelstoactiondexterousarmhand}. Notably, all of these methods rely on perception modalities like depth, tactile sensing, or point cloud inputs, which have gained significant attention due to the relatively small Sim2Real gap they offer. The reduced discrepancy between simulated and real-world data in these modalities has led to remarkable progress, reinforcing the idea that \textit{policies trained on modalities that can be simulated well, can be transferred well.} 
 \begin{figure}[ht]
    \centering
    \includegraphics[width=\linewidth]{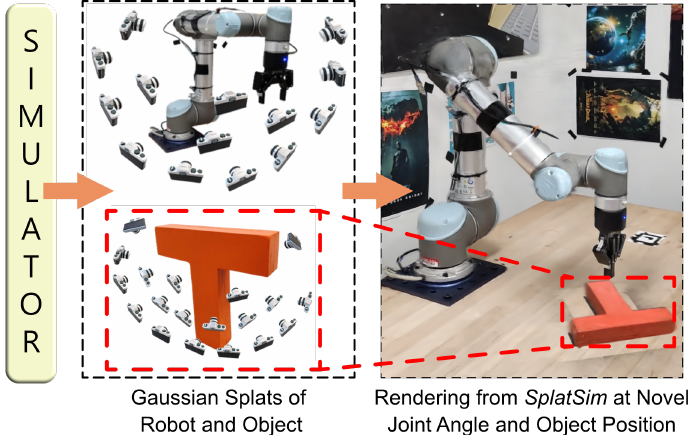}
    \vspace{-15pt}
    \caption{We employ Gaussian Splatting \cite{kerbl3Dgaussians} as the primary rendering primitive within existing simulation environments to generate highly photorealistic synthetic data for robotic manipulation tasks. Our framework retains all the traditional advantages of simulators—including scalability, cost-efficiency, and safety—while enhancing visual realism. Policies trained exclusively on this synthetic data exhibit zero-shot transfer capabilities to real-world scenarios, achieving performance comparable to those trained on real-world datasets.}
    \vspace{-2em}
    \label{fig:teaser}
\end{figure}
 
In contrast, RGB images are rarely used as the primary sensing modality in robot learning applications. over other commonly used modalities in Sim2Real transfer. They capture crucial visual details such as color, texture, lighting, and surface reflectivity, to name a few, which are essential for understanding complex environments. For instance, in a simple task of plucking ripe fruits, color is a key feature for determining ripeness—an inference straightforward in RGB space but difficult and impractical with depth or tactile inputs. Additionally, RGB images are easy to acquire in real-world environments with cameras and align closely with human perception, making them well-suited for interpreting intricate details in dynamic and complex scenes.

But why has it been difficult to deploy policies trained in simulation with RGB information to the real world? The problem lies in the fact that the distribution of images the robot observes in the simulator is very different from the distribution of images it would see in the real world. This makes ``vision Sim2Real an out-of-domain generalization problem"~\cite{robotics_sim2real_yu2024}, a fundamental challenge in machine learning that is still unsolved. For this reason, policies trained on simulated images often struggle when applied to distributions of real-world images. 

In this paper, we propose a systematic and novel method to reduce the Sim2Real gap for RGB images, by leveraging Gaussian Splatting \cite{kerbl3Dgaussians} as a photorealistic render, using existing simulators as the physics backbone. We propose utilizing Gaussian Splatting~\cite{kerbl3Dgaussians} as the primary rendering primitive, replacing traditional mesh-based representations in existing simulators, to significantly improve the photo-realism of rendered scenes. By integrating these renderings of simulated demonstrations with state-of-the-art behavior cloning techniques, we introduce a framework for zero-shot transfer of manipulation policies trained entirely on simulation data, to the real world. Our key contributions are as follows: 
\begin{itemize}
\item We propose a novel and scalable data generation framework, ``\textit{SplatSim}" for manipulation tasks. \textit{SplatSim} is focused predominantly on bridging the vision Sim2Real gap by leveraging photorealistic renderings generated through Gaussian Splatting, replacing traditional mesh representation in the rendering pipeline of the simulator.
  \item We show how to leverage Robot Splat Models and Object Splat Models, along with the simulator as a physics backend, to generate photorealistic trajectories of robot-object interactions. Our method eliminates the need for the real-world data collection to learn these interactions, and relies solely on an initial video of the static scene with the robot. We further demonstrate how these renderings, when combined with simulated demonstrations, can be utilized to generate high-quality synthetic datasets for behavior cloning methods.
  \item We demonstrate the effectiveness of our framework by deploying RGB policies, trained entirely in simulation, to the real world in a zero-shot manner across four tasks, achieving an average success rate of 86.25\%, compared to 97.5\% for policies trained on the real-world data.
\end{itemize}

%% file: sections/2_related_works.tex
\section{Related Works}
\subsection{Sim2real}
Robotics simulation tools like~\cite{coumans2021, mittal2023orbit, mujoco, Aguero-2015-VRC,gu2023maniskill2,maniskill, tao2024maniskill3gpuparallelizedrobotics, mittal2023orbit} have become invaluable in scaling up robot learning due to several advantages including parallelization, cost and time efficiency, and safety. Recent advancements in transferring learned policies from simulation to the real world have demonstrated impressive results, particularly in domains that leverage modalities with a low Sim2Real gap, such as depth, point cloud, proprioception, or tactile feedback. These modalities have enabled robots to perform contact-rich tasks like quadruped locomotion~\cite{agarwal2022legged, hoeller2023anymalparkourlearningagile, 10610200, zhuang2023robot, pmlr-v205-fu23a} and bipedal locomotion~\cite{HumanoidTransformer2023, bipedal_ashish}, dexterous manipulation~\cite{qi2022inhand, 10610532, dextreme, YinHQCW23, lum2024dextrahgpixelstoactiondexterousarmhand}, manipulation of articulated objects~\cite{Eisner-RSS-22, zhang2023fbpp}, among others~\cite{su2024sim2realmanipulationunknownobjects, zhou2023hacman,loquercio2019deep,huang2023dynamic}. 

However, Sim2Real transfer for RGB-based manipulation policies remains challenging. Several prior works \cite{pmlr-v78-james17a, pmlr-v87-matas18a, pieter_abeel, sim2sim} have explored Sim2Real transfer for RGB-based manipulation policies using domain randomization techniques. These approaches, however, often require task-specific tuning and environment engineering, which can be both labor-intensive and difficult to achieve accurately in traditional physics simulators. In contrast, our method eliminates the need for task-specific adjustments and leverages an initial scan of the target deployment environment, significantly simplifying the process. Other existing approaches such as domain adaptation~\cite{rlcyclegan}, often rely on extensive offline data collection of real-world object interactions. Our method requires only an initial video of the static scene without the need for additional real-world data collection.


In this work, we address the challenge of transferring RGB-based policies for manipulation tasks using Gaussian Splatting, which requires rendering complex interactions between objects and the robot.  A work notably related to ours is RialTo~\cite{Villasevil-RSS-24} which uses a Real2Sim2Real approach similar to ours. However, their policy is still trained on point clouds, which requires depth during execution time. In contrast, \textit{SplatSim} only uses RGB images for learning and policy deployment. Another recent work Maniwhere~\cite{yuan2024learning} does large-scale reinforcement learning in simulation and shows generalization to the real world, however, their method still requires depth at test time and cannot work with just RGB images in the real world.

\subsection{Gaussian Splatting for Robotics}
Gaussian Splatting~\cite{kerbl3Dgaussians} is a state-of-the-art rendering technique that models scenes using 3D Gaussian primitives, offering an efficient and photorealistic representation of complex geometries. In contrast to NeRF~\cite{mildenhall2020nerf} and its derivatives~\cite{mueller2022instant, barron2021mipnerf, yu2021pixelnerf}, the explicit, point cloud-like structure of Gaussian Splats enables easier manipulation, which has led to numerous subsequent works focused on dynamic Gaussian Splatting models~\cite{luiten2023dynamic, Wu_2024_CVPR, chen2023periodic, bae2024ed3dgs, xie2023physgaussian}. This explicit nature of Gaussian Splatting has also garnered interest in the robotics community, with recent studies applying it to language-guided manipulation~\cite{shorinwa2024splatmover}, object grasping~\cite{ji2024graspsplats}, and deformable object manipulation~\cite{duisterhof2024deformgs}. The two related works, Embodied Gaussians \cite{abou-chakra2024physically} and RoboStudio \cite{lou2024robogsphysicsconsistentspatialtemporal}, focus on learning from real-world data. Embodied Gaussians~\cite{abou-chakra2024physically}  directly learns a forward model for robot-object interactions, requiring real-world data for each new robot and object, while we offload dynamics to a physics engine and focus on RGB-based policy deployment. RoboStudio~\cite{lou2024robogsphysicsconsistentspatialtemporal} combines simulation with Gaussian Splatting but focuses on system identification, whereas our approach generates synthetic data for real-world deployment using existing simulators. Another closely related work to our method is \cite{quach2024gaussian}, which combines Gaussian Splatting with a simulator, but it is focused on navigation. Unlike manipulation tasks, the agent in their approach does not interact with or manipulate the environment.

%% file: sections/3_method.tex
\begin{figure*}[ht]
    \vspace{0.15cm}
    \centering
    \includegraphics[width=0.85\linewidth]{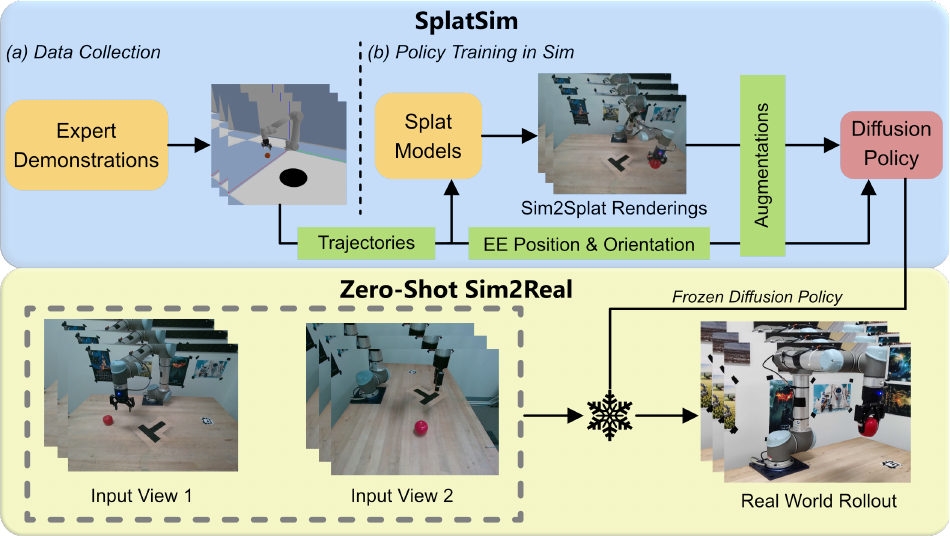}
    \caption{\textbf{Top:} Our proposed SplatSim framework. Expert demonstrations are collected (a) in a physics simulator (PyBullet). In our case, these demonstrations come either from human experts (teleoperation via Gello \cite{wu2023gello}) or through a privileged information-based motion planner. The trajectories from the simulator are then fed to the simulator-aligned splat models of the scene and the object (b). We transform the 3D Gaussians to manipulate the static Gaussian Splat models, as delineated in Sec. \ref{sec:robot_splat_models}, to extract photorealistic renderings of the scene at novel joint and object poses, which serve as the RGB state observations for the diffusion policy. Along with these RGB observations, diffusion policy \cite{chi2023diffusionpolicy} also takes the end effector position and orientation as the input. We augment the end effector states as well. \textbf{Bottom:} Once trained with the sim data, we freeze the policy and directly deploy it to the real-world setting.}
    \label{fig:pipeline}
    \vspace{-1em}
\end{figure*} 
\section{Preliminary}



\subsection{Rigid Body Transformations in Gaussian Splatting}
In Gaussian Splatting, segmented objects within a scene can undergo rigid body transformations, such as translation and rotation, while still maintaining high-quality renderings. Each object, represented by a set of 3D Gaussians, can be transformed using a homogeneous transformation matrix $T$, defined by the rotation \( R \) and translation \( t \). For a 3D Gaussian with mean position \( \mu \) and covariance matrix \( \Sigma \), the transformed position \( \mu' \) and covariance \( \Sigma' \) under the rigid transformation are given by:
\begin{equation}
    \mu' = R \mu + t \label{eq:mu}
\end{equation}
\vspace{-1.9em} 
\begin{equation}
    \Sigma' = R \Sigma R^T \label{eq:sigma}
\end{equation}

Applying these transformations updates the position and orientation of the 3D Gaussians of the object while preserving the corresponding geometric properties. Despite making these changes in the object's configuration, the Gaussian representation enables smooth and accurate renderings. 


\section{Method}
The key premise of our method is that if each rigid body in the Gaussian Splat representation of the real-world scene can be accurately segmented, and its corresponding homogeneous transformation relative to the simulator is identified, then it becomes feasible to render the rigid body in novel poses. The rigid bodies can include links of the robot, links of the gripper, articulated objects, or simple non-deformable objects. By applying this process to all rigid bodies interacting with the robot in simulation, we can generate photorealistic renderings for an entire demonstration trajectory. This approach is analogous to traditional rendering in simulators; however, instead of using mesh primitives, we utilize Gaussian Splats as the underlying representation. This approach allows us to be more effective at capturing the detailed visual fidelity of real-world scenes. 

The following subsections describe our method. We begin by formalizing the problem statement in Sec.~\ref{sec:problem_statement} and notations in Sec.~\ref{sec:notation}. Next, we detail the segmentation and rendering of each robot link at novel joint poses in Sec.~\ref{sec:robot_splat_models}, individual objects at new positions in Sec.~\ref{sec:object_splat_models}, and grippers in Sec.~\ref{sec:articulated_splat_models}. Sec.~\ref{sec:rendering_overall_trajectory} covers the rendering of complete robot-object interaction trajectories, followed by the policy training protocol in Sec.~\ref{sec:policy_training}.



\subsection{Problem Statement}
\label{sec:problem_statement}
We define \( \mathcal{S}_{real} \) as the Gaussian Splat of a real-world scene, captured from multiple RGB viewpoints, including the robot. We also define \( \mathcal{S}^{k}_{obj} \) as the splat of the \( k \)-th object in the scene, captured from multiple viewpoints. Our goal is to use \( \mathcal{S}_{real} \) for generating photorealistic renderings \( {I}^{sim} \) of a robot operating in any simulator (e.g., PyBullet). Then, we can leverage this representation to collect demonstrations using the expert \( \mathcal{E} \) for training RGB-based policies.

The expert \( \mathcal{E} \) generates a trajectory \( \tau_{\mathcal{E}}\) consisting of state-action pairs \(\{(s_1, a_1), \dots, (s_T, a_T)\} \) for a full episode. The state at each time step \( t \) is defined as \( s_t = (q_t, x^1_t, \dots, x^n_t) \), where \( q_t \in \mathbb{R}^m \) denotes the robot's joint angles and \( x^k_t = (p^k_t, R^k_t) \) represents the position \( p^k_t \in \mathbb{R}^3 \) and orientation \( R^k_t \in SO(3) \) of the \( k \)-th object in the scene. The corresponding action \( a_t = (p^e_t, R^e_t) \) refers to the end effector's position \( p^e_t \in \mathbb{R}^3 \) and orientation \( R^e_t \in SO(3) \).

The renderings \( {I}^{sim} \), derived from these simulated states \( s_t \), are used as inputs to train the policy \( \pi_{\mathcal{I}} \). The policy relies solely on real-world RGB images \( {I}^{real} \) at test time.

\subsection{Definitions of Coordinate Frames and Transformations}
\label{sec:notation}
We define several coordinate frames to clarify the relationships between the real-world scene, the simulator, and the splat point clouds. The real-world coordinate frame, denoted as \( \mathcal{F}_{real} \), serves as the primary reference frame. Both the simulator coordinate frame, \( \mathcal{F}_{sim} \), and the real-world robot frame, \( \mathcal{F}_{robot} \), are aligned with \( \mathcal{F}_{real} \). This alignment ensures that the robot's base in the simulator and the real world share the same coordinate system. 

Additionally, the splat coordinate frame, denoted as \( \mathcal{F}_{splat} \), represents the frame of the base of the robot in the Gaussian Splat of the scene \(\mathcal{S}_{real}\). The robot base in the splat point cloud has a different frame from \( \mathcal{F}_{real} \), and we account for this difference by using the transformation matrix \( T^{\mathcal{F}_{splat}}_{\mathcal{F}_{robot}} \).

We also define the \( k \)-th object frame in the simulator, \( \mathcal{F}_{k-obj,sim} \), where objects are initialized in \( \mathcal{SIM} \) at the origin with no rotation. The \( k \)-th object frame in the splat, \( \mathcal{F}_{k-obj,splat} \), represents the object's position and orientation in its Gaussian splat \( \mathcal{S}_{obj} \). The object frames \( \mathcal{F}_{k-obj,sim} \) and \( \mathcal{F}_{k-obj,splat} \) are later aligned during the simulation and splat process using the transformation matrix \( T^{\mathcal{F}_{k-obj,sim}}_{\mathcal{F}_k-{obj,splat}} \).

\begin{figure}[ht]
    \vspace{0.20cm}

    \centering
    \includegraphics[width=\linewidth]{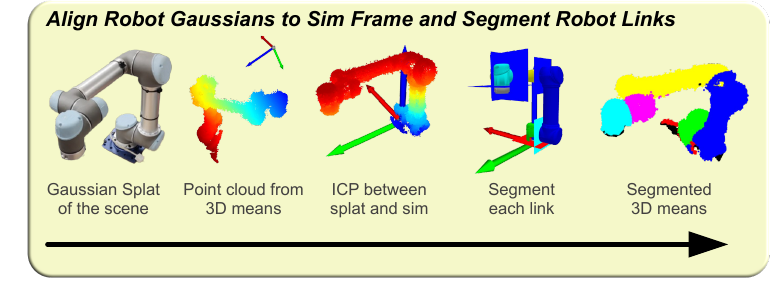}
    \caption{The robot is visualized in a static scene by first creating a Gaussian splat of the scene with the robot in its home position. The robot's point cloud is manually segmented and aligned with the canonical robot frame using the ICP algorithm. Each robot link is then segmented, and forward kinematics transformations are applied, enabling the rendering of the robot at arbitrary joint configurations.}
    \vspace{-6pt}
    \label{fig:robot_splat_models}
\end{figure}

\subsection{Robot Splat Models}
\label{sec:robot_splat_models}
Our method for obtaining robot renderings at novel joint poses is summarized in Fig.~\ref{fig:robot_splat_models}. It follows a three-step approach: 

\subsubsection{Alignment of Gaussian Splat Robot Frame to the Simulator Frame} In order to combine the Gaussian Splat representation \( \mathcal{S}_{real} \) with the simulator, we first manually segment out the 3D Gaussians associated with the robot. The means of these 3D Gaussians form a point cloud which is aligned with the ground truth point cloud obtained from the simulator. To achieve this, we use the Iterative Closest Point (ICP) algorithm, which produces the desired transformation \(T^{\mathcal{F}_{splat}}_{\mathcal{F}_{robot}}\).

\subsubsection{Segmentation of the Robot Links}
To associate the 3D Gaussians with their respective links in \( \mathcal{S}_{real} \), we leverage the ground truth bounding boxes of the robot's links, provided by its CAD model. This method allows us to isolate the 3D Gaussians corresponding to each link in the real-world scene, denoted as \( \mathcal{S}^{l}_{real} \), where \( l \) refers to the \( l \)-th link of the robot.

\subsubsection{Forward Kinematics Transformation}
Once we have the 3D Gaussians for individual links and the frames aligned, we can use the robot's forward kinematics to get the robot pose at arbitrary joint angles \(q_t \in s_t\), given by the simulator. In this work, we use the forward kinematics routine from PyBullet to get the Transformation $T^{l}_{fk}$ for link $l$ in the robot's canonical frame \( \mathcal{F}_{sim} \). The transformation of the 3D Gaussians can be calculated as : 
\begin{align}
\label{eq:robotrendering}
\textstyle T = (T^{\mathcal{F}_{splat}}_{\mathcal{F}_{robot}})^{-1} \cdot T^{l}_{fk} \cdot T^{\mathcal{F}_{splat}}_{\mathcal{F}_{robot}}
 \end{align}
 where $T^{\mathcal{F}_{splat}}_{\mathcal{F}_{robot}}$ is the transformation matrix to get the robot from splat frame to the simulation frame. Once the transformation for each link is calculated, we use Eq.~\ref{eq:mu} and Eq.~\ref{eq:sigma} to transform the 3D Gaussians related to individual links of the robot. The robot at novel poses is then rendered by the standard Gaussian Splatting rendering framework~\cite{kerbl3Dgaussians}.

\subsection{Object Splat Models}
\label{sec:object_splat_models}
Similar to the robot rendering, we use ICP to align each object's 3D Gaussians \( \mathcal{S}^{k}_{obj} \) to its simulated ground truth point cloud. In this way, we get the transformation \(T_{\mathcal{F}_{k-obj,sim}}^{\mathcal{F}_{k-obj,splat}} \), which transforms the splat in \( \mathcal{S}^{k}_{obj} \) frame to simulator frame. Given the position \(p^{k}_t \in s_t\) and orientation \(R^{k}_t \in s_t\) can be used to calculate the transformation of object \(T_{\text{fk}}^{k-obj}\) from its original simulator frame \( \mathcal{F}_{k-obj,sim} \). Using \(T_{\text{fk}}^{k-obj}\) we can get the object's 3D Gaussians in \( \mathcal{S}_{real} \) frame with the transformation : 
\begin{align}
\label{eq:objectrendering}
\textstyle T =  (T^{\mathcal{F}_{splat}}_{\mathcal{F}_{robot}})^{-1} \cdot T^{k-obj}_{fk} \cdot  T_{\mathcal{F}_{k-obj,sim}}^{\mathcal{F}_{k-obj,splat}}
 \end{align}

 Once the transformation for the object is calculated, we can again use Eq.~\ref{eq:mu} and Eq.~\ref{eq:sigma} to transform the 3D Gaussians related to the object. Then we use the Gaussian Splatting rendering framework~\cite{kerbl3Dgaussians} to render the object at its new position and orientation.

\begin{figure}[ht]
    \centering
    \includegraphics[width=\linewidth]{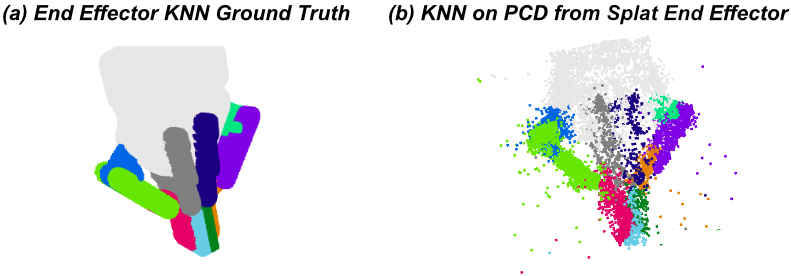}
    \caption{We use a KNN-based classifier for segmenting links for articulated objects like parallel jaw grippers. We train a KNN model with the ground truth point labeling from the URDF model of the end effector.}
    \label{fig:knn_splat}
    \vspace{-15pt}
    
\end{figure}
 
\subsection{Articulated Object}
\label{sec:articulated_splat_models}
While CAD axis-aligned bounding boxes allow straightforward segmentation of robot links, certain objects, such as parallel jaw grippers, present challenges due to their misalignment with standard axes, that is, the gripper links are not neatly segmented out by just using bounding boxes in the 3D space. To address this, we employ a ground truth K-Nearest Neighbour (KNN) classifier trained on labeled simulator point clouds as in Fig.~\ref{fig:knn_splat} (a), which enables inference of the link class for each 3D Gaussian in the aligned splat as shown in Fig.~\ref{fig:knn_splat} (b).

\subsection{Rendering Simulated Trajectories using SplatSim}
\label{sec:rendering_overall_trajectory}
Now that we are able to render individual rigid bodies in the scene, we can use this to represent any simulated trajectory \( \tau_{\mathcal{E}}\) with photorealistic accuracy. We use these state-based transformations along with methods described in Sec.~\ref{sec:robot_splat_models}, \ref{sec:object_splat_models} to get the demonstration for our policy to learn from \( \tau_{\mathcal{G}} = \{(I^{sim}_1, a_1), (I^{sim}_2, a_2), \dots, (I^{sim}_T, a_T)\} \). This data is used by policy to predict actions from the synthetically generated images.

\begin{table*}[ht]
    \vspace{0.25cm}

    \centering
    \renewcommand{\arraystretch}{1.3} 
    \begin{tabular}{cccc|cc}
        \hline
        \multirow{2}{*}{\textbf{Task}} & \multicolumn{3}{c}{\textbf{Successful Trials}}  & \multicolumn{2}{c}{\textbf{Human Effort to Collect Data}} \\
        & \multicolumn{3}{c}{(Out of 40 Trials)} & \multicolumn{2}{c}{\textbf(hours)}\\
        \cline{2-4}\cline{5-6}
         & Sim2Sim & Real2Real & Sim2Real \textbf{(SplatSim)}& Simulator & Real World \\
        \hline
        T-Push & 100\% & 100\% & 90\% & 3.0 & 3.5 \\
        Pick-Up-Apple & 100\% & 100\% & 95\% & 0.0\(^{\ast}\) & 3.5 \\
        Orange-On-Plate & 97.5\% & 95\% & 90\% & 0.0\(^{\ast}\) & 6.0 \\
        Assembly & 85\% & 90\% & 70\% & 0.0\(^{\ast}\) & 7.5\\
        \cline{1-6}
        Total & 95.62\% & 97.5\% & \textbf{86.25}\% & 3.0 & 20.5 \\
        \hline
        \multicolumn{6}{l}{$^{\ast}$ Automated process} \\
    \end{tabular}
    
    \caption{Comparison of task success rates and data collection times across various manipulation tasks. Our policies trained solely on synthetic data achieve an 86.25\% zero-shot Sim2Real performance, comparable to those trained on real-world data. By leveraging the automation capabilities of simulators, we significantly reduce the human effort required for data generation.}
    
    \label{tab:main_result}
\end{table*}

\subsection{Policy Training and Deployment}
\label{sec:policy_training}
For learning from the generated demonstrations \( \tau_{\mathcal{G}}\) in the simulator, we employ Diffusion Policy~\cite{chi2023diffusionpolicy, chi2024diffusionpolicy}, which is the state of the art for behavior cloning. Although our method significantly mitigates the vision Sim2Real gap, discrepancies between the simulated and real-world environments remain. For instance, simulated scenes lack shadows, and rigid body assumptions can lead to improper rendering of flexible components such as robot cables. To address these issues, we incorporate image augmentations similar to~\cite{nerf2real} during policy training, which includes adding gaussian noise, random erasing and adjusting brightness and contrast of the image. These augmentations notably enhance the robustness of the policy and improve its performance during real-world deployment.

%% file: sections/4_experiments.tex
    \section{Experiments}    
   To evaluate the effectiveness of our framework in bridging the Sim2Real gap for RGB-based manipulation tasks, we conducted extensive experiments across four real-world manipulation tasks. We begin by detailing the data collection process in both the simulator and real-world environments. We then compare the performance of policies trained on our synthetic data with Real2Real policies—those trained on real-world data and deployed in real-world environments. This comparison demonstrates the high fidelity of our synthetic data, showing that policies trained within our framework can be deployed to real-world tasks without fine-tuning. Additionally, we assess Sim2Sim performance by training and evaluating policies entirely within the \textit{SplatSim}, allowing us to quantify the degradation in performance during Sim2Real transfer. Lastly, we investigate the effects of data augmentation on the transfer process and evaluate the visual fidelity of the photorealistic renderings generated by the \textit{SplatSim} framework. 

\subsection{Demonstrations in the Real World and Simulation}
In the real world, demonstrations for each task were manually collected by a human expert. In contrast, the simulator streamlines this process by employing privileged information-based motion planners, which automatically generate data using privileged information, such as the position and orientation of each rigid body in the scene. The simulator not only reduces effort by automating resets between demonstrations when a human expert is involved but more importantly, it leverages motion planners that eliminate the need for human intervention entirely. This enables the generation of large-scale, high-quality demonstration datasets with minimal manual input. As a result, the simulator drastically reduces the time and effort required for data collection. As shown in Table~\ref{tab:main_result}, while real-world demonstration collection required about 20.5 hours, the same tasks were completed in just 3 hours in the simulator, underscoring the efficiency and scalability of our approach.

\subsection{Zero-Shot Policy Deployment Results}

We evaluate the zero-shot deployment of our policies across four contact-rich real-world tasks, using task success rate as the primary metric. As shown in Table~\ref{tab:main_result}, our method achieves an average success rate of 86.25\% for zero-shot Sim2Real transfer, compared to 97.5\% for policies trained directly on real-world data, highlighting the effectiveness of our approach. All experiments were conducted using a UR5 robot equipped with a Robotiq 2F-85 gripper and 2 Intel Realsense D455 cameras~\cite{8014901} with deployment on an NVIDIA RTX 3080Ti GPU for the Diffusion Policy \cite{chi2023diffusionpolicy}.
\label{zero_shot_deployment}

\begin{figure*}[ht]
    \vspace{0.15cm}

    \centering
    \includegraphics[width=\linewidth]
    {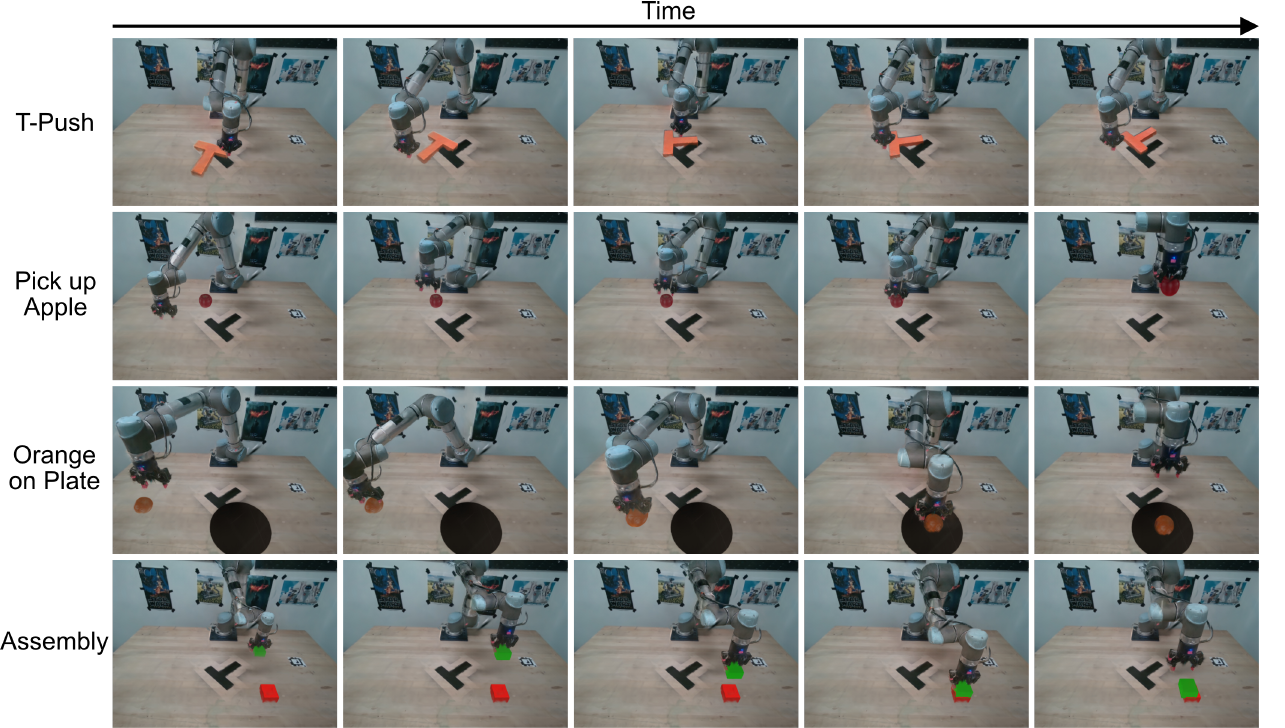}
    \caption{SplatSim Rollout: enderings from our SplatSim framework across four different manipulation tasks. }
    \label{fig:qualitative}
    
\end{figure*}


\subsubsection{T-Push Task}
\label{tpush_task}
The T-Push task, popularized by Diffusion Policy \cite{chi2023diffusionpolicy}, captures the dynamics of non-prehensile manipulation, which involves controlling both object motion and contact forces. For training, a human expert collected 160 demonstrations in simulation using the Gello teleoperation \cite{wu2023gello}. While testing, the robot started from a random location and achieved a 90\% success rate (36/40 trials) in zero-shot Sim2Real transfer as shown in Table \ref{tab:main_result}. This result shows the effectiveness of our framework in handling the dynamics of pushing without fine-tuning on real-world demonstrations. Additionally, the performance of our method is comparable to Real2Real (40/40) and Sim2Sim (40/40).

\subsubsection{Pick-Up-Apple Task}
The Pick-Up-Apple task involves grasping and manipulating the full pose of an object (i.e., position and orientation) in 3D. This task was designed to evaluate the robot's grasping capabilities when trained using our simulated renderings. A motion planner, leveraging privileged state information from the simulator (accurate position and orientation of each rigid body in the scene), generated 400 demonstrations with randomized end-effector positions and orientations. During real-world trials, our policy achieved a 95\% success rate (38/40 trials) in zero-shot Sim2Real transfer, as shown in Table \ref{tab:main_result}.
\label{apple_pick}


\subsubsection{Orange on Plate Task}
In this task the robot has to pick up an orange and place it on a plate. In simulation, a motion planner with access to privileged information, generated 400 demonstrations. The end-effector position and initial gripper state were randomized during training. During testing, the robot always started from a home position. We achieved a 90\% success rate (36/40 trials) in zero-shot Sim2Real transfer.
\label{orange}

\subsubsection{Assembly Task}
In this task the robot has to put a cuboid block on top of another cuboid. The robot starts at the home position with the green cube already grasped and has to place it on top of the red cube.  The task is particularly tough since the robot has to make a precise placement otherwise the cube will fall and will lead to a failure case. Our Sim2Real policy achieved a performance of 70\% (28/40 trials) on this task, compared to 95\% on Sim2Sim and 90\% on Real2Real. 
 \label{assembly}


\subsection{Quantifying Robot Renderings}
We quantitatively evaluate the accuracy of rendered robot images at various joint configurations by comparing them with the real-world images. We assess the quality of the robot's renderings across 300 different robot joint angles. To measure the similarity between the rendered and real-world images, we employ two metrics commonly used in image rendering assessment: Peak Signal-to-Noise Ratio (PSNR) and Structural Similarity Index Measure (SSIM). Despite the variations in joint configurations, the renderings achieve an average PSNR of 22.62 and an SSIM of 0.7845, indicating that the simulated images closely approximates the visual quality of the real-world RGB observations.



\subsection{Effect of Augmentations}
To quantify the impact of data augmentations on the Sim2Real performance of our policy, we conducted experiments comparing policies trained with and without augmentations. While the Diffusion-Policy performs effectively without augmentations in consistent environments (e.g., Sim2Sim or Real2Real scenarios), transferring a policy trained in simulation to the real world introduces domain shifts that necessitate additional robustness as the renderings can't capture dynamic details like changing reflections and shadows. We incorporated augmentations such as random noise addition, Color Jitter, and random erasing during training to address these shifts. These augmentations improve the performance of the policy from 21\% to 86.25\% across four tasks in Sec. \ref{zero_shot_deployment}.


%% file: sections/5_conclusion.tex
\section{Conclusion}
In this work, we tackled the challenge of reducing the Sim2Real gap for RGB-based manipulation policies by leveraging Gaussian Splatting as a photorealistic rendering technique, integrated with existing simulators for physics-based interactions. Our framework enables zero-shot transfer of RGB-based manipulation policies trained in simulation to real-world environments. While our framework advances the current state-of-the-art, it is still limited to rigid body manipulation and cannot handle complex objects such as cloth, liquids, or plants. In the future, our plan is to combine our framework with reinforcement learning-based methods to acquire more dynamic skills. We will also further improve our system to train and deploy robots in highly complex and contact-rich tasks in the real world. Specifically, agricultural tasks such as pruning and harvesting, which require data that is challenging to obtain under field conditions, could greatly benefit from our proposed method.